\PassOptionsToPackage{table}{xcolor}
\documentclass{article} 
\usepackage[table]{xcolor}
\usepackage{icomp2024_conference,times}

\usepackage{amsmath,amsfonts,bm}

\def\eqref#1{equation~\ref{#1}}

\def\1{\bm{1}}

\DeclareMathAlphabet{\mathsfit}{\encodingdefault}{\sfdefault}{m}{sl}
\SetMathAlphabet{\mathsfit}{bold}{\encodingdefault}{\sfdefault}{bx}{n}

\newcommand{\E}{\mathbb{E}}

\newcommand{\R}{\mathbb{R}}

\usepackage{xcolor}
\usepackage{algorithm}
\usepackage{algorithmic}
\usepackage{amsmath}
\usepackage{amssymb}
\usepackage{hyperref}
\usepackage{url}
\usepackage{svg}
\usepackage{booktabs}
\usepackage{pifont}
\usepackage{dot2texi}
\usepackage{tikz}
\usetikzlibrary{shapes,arrows}
\usepackage{graphicx}
\usepackage{url}
\definecolor{CustomColor}{HTML}{D5F5E3}

\title{AdLoCo: adaptive batching significantly improves communications efficiency and convergence for Large Language Models}

\icompfinalcopy

\author{Nikolay Kutuzov\thanks{These authors contributed equally to this work.} \\
Department Mathematical Foundations of Control \\
Moscow Institute of Physics and Technology \\
Moscow, Russia \\
\texttt{kutuzov.nv@phystech.edu} \\
\And 
Makar Baderko\footnotemark[1]\\
The School of the Center for Teacher Excellence \\
Moscow, Russia \\
\texttt{makarbaderko@gmail.com}
\And Stepan Kulibaba\footnotemark[1], \ Artem Dzhalilov \\
Research Center of the Artificial Intelligence \\ 
Institute, Innopolis University, \\
Innopolis, Russia \\
\texttt{kulibabast@gmail.com} \\
\texttt{artem.dzhalilov@gmail.com}
\And 
{Daniel Bobrov} \\
Sirius University of Science and Technology \\
Sirius, Russia \\
\texttt{bobrovdanel@gmail.com}
\And 
Maxim Mashtaler \\
Department Mathematical Foundations of Control \\
Moscow Institute of Physics and Technology \\
Moscow, Russia \\
\texttt{mashtaler.mk@mipt.ru} \\
\texttt{mashtaler.mk@phystech.edu} \\
\And Alexander Gasnikov \\
Innopolis University, MIPT \& Steklov Institute, Russia \\
\texttt{gasnikov@yandex.ru}
}

\usepackage{amsmath,amsthm}

\begin{document}

\newtheorem{lemma}{Lemma}
\newtheorem{theorem}{Theorem}

\maketitle

\begin{abstract}
Scaling distributed training of Large Language Models (LLMs) requires not only algorithmic advances but also efficient utilization of heterogeneous hardware resources. While existing methods such as DiLoCo have demonstrated promising results, they often fail to fully exploit computational clusters under dynamic workloads. To address this limitation, we propose a three-stage method that combines Multi-Instance Training (MIT), Adaptive Batched DiLoCo, and switch mode mechanism. MIT allows individual nodes to run multiple lightweight training streams with different model instances in parallel and merge them to combine knowledge, increasing throughput and reducing idle time. Adaptive Batched DiLoCo dynamically adjusts local batch sizes to balance computation and communication, substantially lowering synchronization delays. Switch mode further stabilizes training by seamlessly introducing gradient accumulation once adaptive batch sizes grow beyond hardware-friendly limits. Together, these innovations improve both convergence speed and system efficiency. We also provide a theoretical estimate of the number of communications required for the full convergence of a model trained using our method.
\end{abstract}

\section{Introduction}
The rapid ascent of Large Language Models (LLMs) has turned them into the central tool of modern machine learning, yet their training remains an exceptionally resource-intensive process. Scaling  Large Language Models (LLMs) across distributed clusters requires not only algorithmic innovation but also strategies for efficient utilization of heterogeneous hardware. While recent approaches such as DiLoCo (\cite{douillard2024dilocodistributedlowcommunicationtraining} and classical algorithms such as LocalSGD (\cite{localsgd}) have demonstrated the potential of distributed optimization, they often utilize computational resources inefficiently under dynamic workloads, leaving efficiency gaps.

In this work, we present a three-stage methodology designed to close these gaps and push the limits of distributed training. Our approach combines Multi-Instance Training (MIT), Adaptive Batching for DiLoCo, and a switch mode mechanism. MIT enables each compute node to host multiple lightweight training streams on different model instances, which are periodically merged to exchange knowledge. This not only maximizes hardware throughput but also reduces idle time in multi-node environments. Moreover, adaptive batching dynamically adjusts local batch sizes to balance computational workloads and provide for more efficient use of available resources, while also significantly diminishing synchronization overhead. Our adaptive batching framework is built up on the results shown in AdAdaGrad (\cite{adadagrad}). Given the foregoing, the natural interaction between MIT and adaptive batching yields a compounding benefit: as we approach the solution, fewer parallel instances remain active, but every instance uses a larger batch size, so the training process becomes more communication-efficient.

To complement these techniques, the switch mode mechanism introduces gradient accumulation once adaptive batch sizes surpass hardware-friendly thresholds, 
ensuring robustness under batch-size limitations. Together, the three introduced components establish an effective synergy: multi-instance flexibility, memory-aware batching, and stability-preserving switching.

We further support our methodology with a theoretical upper bound on the number of inter-instance communications required for convergence, yielding a rigorous understanding of its communication complexity. Our method achieves faster time-to-target perplexity compared to widely used distributed training baselines on industry standard benchmarks, while preserving convergence quality. These results highlight not only the practical value of our approach, but also its potential as a foundation for the next generation of scalable distributed LLM training. To facilitate reproducibility, we provide the implementation at \url{https://github.com/funmagster/AdLoCo}.

\begin{figure}[h]
    \centering
    \includegraphics[width=0.7\textwidth]{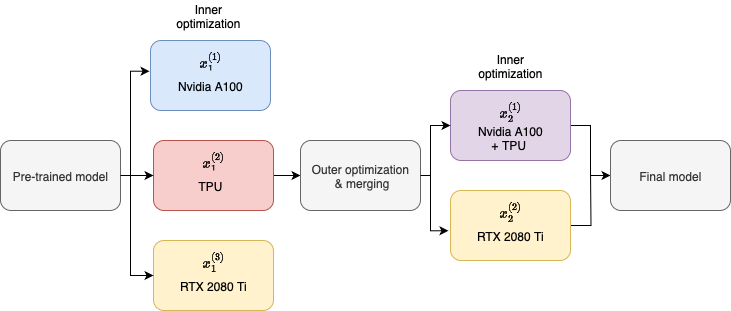}
    \label{fig:intro_plot}
\end{figure}

\section{Preliminaries}

\subsection{Notation}  

In this text, we denote the model's parameters of the $i$-th instance at iteration  $k$ as $x_k^{(i)}$. The dataset is partitioned among $M$ trainers belonging to $\mathbb{T}$, where the data shard $\mathcal{D}_i$ corresponds to the $i$-th trainer. We use $f(x_k^{(i)}; \xi)$ to denote the loss function  on a data sample $\xi$, with model of the $i$-th trainer at state $x_k^{(i)}$.

Let $B_k$ denote the mini-batch of samples used at iteration $k$ with batch size $b_k$, and define the mini-batch gradient as:
\begin{equation}
g_k = \nabla F_{B_k}(x_k) := \frac{1}{b_k} \sum_{\xi \in B_k} \nabla f(x_k; \xi). \label{eq:1}
\end{equation}

\subsection{Problem setting}

We consider the classical stochastic optimization problem in the distributed setting:
\begin{equation} \label{eq:opt}
\min_{x \in \mathbb{R}^d} F(x) := \mathbb{E}_{\xi \sim \mathcal{D}} [f(x; \xi)],
\end{equation}
where $f$ is a smooth, possibly non-convex loss function, and $\mathcal{D}$ is the underlying data distribution.

In the explored learning setting, the global objective \eqref{eq:opt} is optimized across training instances $\mathbb{T}$, each holding a local data shard $\mathcal{D}_i$. Each trainer performs local updates using its own data and periodically synchronizes the local model's  parameters with other trainers to reduce communication costs.

\section{Related work}

The problem of efficient distributed optimization has received considerable attention in recent years, particularly in the context of large-scale deep learning. Classical distributed methods, such as LocalSGD \cite{localsgd}, reduce communication costs by allowing local computation between synchronization rounds, but they do not incorporate adaptive mechanisms to adjust training dynamics. Subsequent approaches like DiLoCo \cite{douillard2024dilocodistributedlowcommunicationtraining} introduced hierarchical communication patterns, further lowering synchronization overhead, yet they similarly lack adaptivity in batch sizing or coordination policies. On a different line of research, adaptive optimization methods such as AdAdaGrad \cite{adadagrad} explored variance-aware batch size adjustment strategies to improve convergence and computational efficiency. However, these methods are typically designed for single-worker or data-parallel setups without addressing distributed trainer coordination.  

\subsection{LocalSGD}

LocalSGD (\cite{localsgd}) is a commonly used distributed learning algorithm, which introduced the concept of federated periodic averaging.

With \ref{eq:opt} as the optimization objective, $M$ parallel workers and data shards $\mathcal{D}_i$, at each step the $i$-th worker performs a local SGD step:

\begin{equation} \label{eq:11}
    x_{k+1}^{(i)} = x_k^{(i)} - \eta_k \nabla f(x_k^{(i)};\xi_k^{(i)})
\end{equation}

Instead of synchronizing after every step, as commonly seen in classic distributed learning algorithms, \cite{localsgd} proposed averaging the gradients only in every $H$ steps. Specifically, if the next step is the averaging step ($k + 1 \ \text{mod} \ H = 0$), we perform the synchronization:

\begin{equation}
    x_{k+1}^{(i)} = \frac{1}{M} \sum_{i=1}^M x_k^{(i)}
\end{equation}

After synchronization during step $t + 1$, the weights are changed according to \ref{eq:11} as usual, therefore, resulting in the complete LocalSGD's weights update algorithm: 

\begin{equation}
x_{k+1}^{(i)} =
\begin{cases}
\displaystyle x_{k}^{(i)} - \eta_k\,\nabla f\!(x_{k}^{(i)};\,\xi_{k}^{(i)})\,, & \text{if } k + 1 \ \text{mod} \ H \neq 0, \\[1em]
\displaystyle \frac{1}{M} \sum_{j=1}^M \bigl(x_{k}^{(j)} - \eta_k\,\nabla f\!(x_{k}^{(j)};\,\xi_{k}^{(j)})\bigr), & \text{if } k + 1 \ \text{mod} \ H = 0
\end{cases}
\end{equation}

\subsection{DiLoCo}

DiLoCo is a distributed optimization algorithm, specifically, a variant of federated averaging. We assume that we have a base model  (e.g., a transformer), that we want to train using a multi-instance procedure with its parameters denoted as $x$. In context of language modelling a series of tokens serves as the input, while the target is the input sequence that has been shifted by one token. Due to the multi-instance nature of the DiLoCo (\cite{douillard2024dilocodistributedlowcommunicationtraining})
algorithm the dataset is splitted into $n$ data shards, where the $i$-th data shard is denoted as $\mathcal{D}_i$.

The algorithm consists of two optimization processes denoted as outer and inner optimization, respectively. During $H$ steps of the inner optimization stage each worker performs independently, calculating the gradients on its own data shard as seen in standard single-instance machine learning scenarios. During each of the $T$ steps of the outer optimization stage, the gradients from each worker are gathered and averaged in the outer gradient. Therefore, every worker trains $H \times T$ total training steps during the procedure.

\subsection{AdAdaGrad}

AdAdaGrad (\cite{adadagrad}) is an optimization algorithm, which combines AdaGrad's (\cite{adagrad}) learning rate adaptiveness with adaptive batch size scheduling, yielding better convergence $\&$ more efficient GPU usage. The paper considers the standard optimization problem of minimizing the loss function (\ref{eq:opt}) over the dataset.

The minimized function is assumed to be smooth and, possibly, non-convex. The stochastic gradient $f(x; \xi)$ will be assumed to be an unbiased estimator for the true gradient $F(x)$, that is, $\mathbb{E}_\xi[\nabla f(x; \xi)] = \nabla F(x)$. 

\cite{adadagrad} explore two different adaptive sampling algorithms (norm test and augmented inner product test).

\subsubsection{Norm test}

\cite{norm_test} has shown the norm test to be an optimal strategy: 

If $f(x,\xi)$ is a convex and continuously differentiable function for any $\xi \in \mathcal{Z}$, then $- \nabla F_B(x)$ is a descent direction for $F$ at $x \in \mathbb{R}^d$ if there exists $\eta \in [0, 1)$ such that

\begin{equation} \label{eq:3}
\delta_B(x) := \left\| \nabla F_B(x) - \nabla F(x) \right\| \leq \eta \left\| \nabla F(x) \right\|
\end{equation}

The used in practice approximate norm test condition can be viewed as:

\begin{equation} \label{eq:4}
    \mathbb{E}\!\Big[\|g_k - \nabla F(x_k)\|^2\Big] \leq \eta^2 \|\nabla F(x_k)\|^2
\end{equation}

Under the assumption of knowledge of the stochastic gradient variance

\begin{equation}
    \sigma_{B_k}^2 = \operatorname{Var}_{i \in B_k} \left( \nabla f(x_k; \xi_i) \right)
\end{equation}

the equation \ref{eq:4} can be transformed into 

\begin{equation}
    b_k \ge \frac{\sigma_{B_k}^2}{\eta^2 \|\nabla F(x_k)\|^2}
\end{equation}

therefore yielding the expression for the batch size at iteration $k + 1$:

\begin{equation}
b_{k+1} = \left\lceil \frac{\sigma_{B_k}^2 }{ \eta^2 \left\| \nabla F_{B_k}(x_k) \right\|^2 } \right\rceil.
\end{equation}

\subsubsection{Inner product test}

\cite{inner_product_test} noted that the aforementioned norm test often leads to a rapid increase in batch size, which could reduce the efficiency of adaptive batching, and proposed the inner product test as an alternative. This test further controls the variance $\sigma_{B_k}^2$ to ensure that $-\nabla F_{Bk}(x_k)$ is a descent direction \textit{with high probability}. The condition \ref{eq:4} transforms into (for an existing $\vartheta > 0$): 

\begin{equation}
\frac{1}{b_k} \mathbb{E}_k \left[ \left( \left\langle \nabla f(x_k; \xi_i), \nabla F(x_k) \right\rangle - \left\| \nabla F(x_k) \right\|^2 \right)^2 \right] \leq \vartheta^2 \left\| \nabla F(x_k) \right\|^4
\end{equation}

which can after equivalent for norm-transform transformations lead to the equation for the batch size at iteration $k + 1$: 

\begin{equation}
b_{k+1} = \left\lceil 
\frac{\mathrm{Var}_{i \in \mathcal{B}_k} \left( \langle \nabla f(x_k; \xi_i), \nabla F_{\mathcal{B}_k}(x_k) \rangle \right)}
{\vartheta^2 \left\| \nabla F_{\mathcal{B}_k}(x_k) \right\|^4}
\right\rceil
\end{equation}

The \cite{adadagrad} paper also mentions the impracticality of the augmented inner product test values computation due to the fact that (near) orthogonality of $\nabla F_{B_k} (x_k)$ and $\nabla F (x_k)$ has not been observed in practice as mentioned in \cite{inner_product_test}. We confirmed this statement in practice, since a $10^7$ order difference was observed between the statistics. For reference, the augmented inner product test utilizes the following adaptive batching strategy:

\begin{equation}
b_{k+1}^{'} = \left\lceil 
\max \left\{b_{k+1},\ 
\frac{\mathrm{Var}_{i \in \mathcal{B}_k} \left( 
\nabla f(x_k; \xi_i) - 
\frac{\langle \nabla f(x_k; \xi_i), \nabla F_{\mathcal{B}_k}(x_k) \rangle}
{\left\| \nabla F_{\mathcal{B}_k}(x_k) \right\|^2}
\nabla F_{\mathcal{B}_k}(x_k)
\right)}
{\nu^2 \left\| \nabla F_{\mathcal{B}_k}(x_k) \right\|^2}
\right\}
\right\rceil.
\end{equation}

\section{AdLoCo}
In this section we present AdLoCo, which augments DiLoCo with adaptive batching, trainer merging, and a switch policy for gradient accumulation. Full algorithm is in the appendix A~\ref{alg:adloco}.


\subsection{Multi-instance training}
Multi-instance training extends distributed optimization by maintaining a collection of $n$ trainers with identical architectures and independent initializations. Each trainer is associated with a random subset $\mathcal{D}^{(i)} \subseteq \mathcal{D}$ of the global dataset. Independent sampling, combined with stochastic optimization noise, generates diverse optimization trajectories, enhances coverage of the loss landscape, and helps prevent premature convergence to suboptimal basins. Periodic consolidation via merging operations combines complementary information and progressively reduces redundancy among trainers.

\subsubsection{Trainer merger}

A trainer is denoted $T_k^{(i)} \in \mathbb{T}$ and may span multiple GPUs. Let $x_k^{(i)}$ and $b_k^{(i)}$ be its parameters and batch size at step $k$, and let $\mathcal{D}^{(i)}$ be the (possibly intersecting) random data subset assigned to trainer $i$.  
Merging is defined over a subset $\mathbb{M}_k \subseteq \mathbb{T}$: these trainers are replaced by a single representative whose parameters are obtained via a weighted average proportional to their batch sizes,
\[
    x^{(i)}_{k+1} =
    \begin{cases}
        x^{(i)}_{k}, & i \notin \mathbb{M}_k,\\[0.6em]
        \displaystyle \frac{\sum_{j \in \mathbb{M}_k} b_k^{(j)} x_k^{(j)}}{\sum_{j \in \mathbb{M}_k} b_k^{(j)}}, & i \in \mathbb{M}_k,
    \end{cases}
\]
after which only one trainer from $\mathbb{M}_k$ is retained and $\mathbb{T}$ is updated accordingly. This procedure yields a controlled contraction of the ensemble while preserving accumulated signal.

\subsubsection{Merging policy}

Under the AdLoCo policy, trainers with the smallest requested batch sizes are selected for merging. Small requested batches are treated as a proxy for slower progress toward the large-batch, low-variance regime. Consolidating these trainers preserves their information within a stronger representative and reallocates computation toward trajectories exhibiting more advanced optimization dynamics.






\subsection{Mode switching}
A practical limitation of adaptive batching is the maximum batch size that can fit into GPU memory, denoted as $\texttt{max\_batch}$. A common workaround is to use gradient accumulation once the requested batch size exceeds $\texttt{max\_batch}$. However, enabling accumulation too early can unnecessarily increase training variance and slow down optimization, since the effective batch size remains moderate while the update frequency is reduced.

To address this, we introduce the \emph{SwitchMode} strategy: gradient accumulation is activated only when the requested batch size surpasses $n \times \texttt{max\_batch}$. The intuition is that, in this regime, the statistical benefits of larger batches outweigh the cost of less frequent parameter updates, making accumulation efficient. In contrast, when the requested batch is only slightly above $\texttt{max\_batch}$, it is preferable to keep standard updates, avoiding the overhead of accumulation while still maintaining stable optimization dynamics. In our experiments, we set $n=2$, which provided a good balance between memory efficiency and convergence speed.

\section{Theoretical Estimate of Communication Complexity}
    
\begin{theorem}[Batch Size Estimate]
For the proposed methodology, using SGD as outer optimizer, the expected batch size at the $k$-th iteration satisfies the lower bound
\[
\mathbb{E}[b_k] = \Omega \left(\frac{k\sigma^2}{\eta^2 L(HM+\eta^2)(F(x_0) - F(x^*))}\right).
\]
\end{theorem}

\begin{theorem}[Communication Complexity]
The expected total number of communications between instances, using SGD as outer optimizer, during training after $N$ gradient accumulation iterations is bounded by
\[
\mathbb{E}[C(N)] = \mathcal{O}\left(\frac{b_{\max}\eta^2 L(1 + \eta^2)(F(x_0)-F(x^*))}{\sigma^2}\ln{N}\right).
\]
\end{theorem}
Proofs of the theorems can be found in the appendix. 

\section{Experiments}

\subsection{Setup}

During training, a single Nvidia A100 GPU with 80 gigabytes of VRAM was used for both DiLoCo and AdLoCo simulations. A cluster of multiple GPUs was simulated on the card using threading, so that the processes were executed independently and asynchronously. During our experiments, we have simulated 4 different GPUs with $80 \ / \ 4 \ = \ 20$ gigabytes of VRAM on each simulated instance.

We used a MicroLlama model (\cite{wang2024microllama}) as a LLM of our choice, since due to VRAM limitations, we could not use the same model as in the original \cite{douillard2024dilocodistributedlowcommunicationtraining} paper.

We pre-trained the model on the English subset of the C4 corpus.
The model was initialized from the \texttt{keeeeenw/MicroLLaMA} checkpoint and trained for 2400 steps with a per-device batch size of 100 and gradient accumulation over 5 steps. 
We used AdamW optimization with a learning rate of $4 \times 10^{-4}$ and weight decay of 0.1. 
The training set contained approximately 3.6M tokenized examples, while evaluation was performed every 10 steps on a held-out validation shard. 
All experiments were conducted with mixed-precision training (FP16) on an NVIDIA A100 GPU.

\subsection{Results}

\begin{figure}[ht]
\centering
\begin{minipage}{0.48\textwidth}
  \centering
  \includegraphics[width=\linewidth]{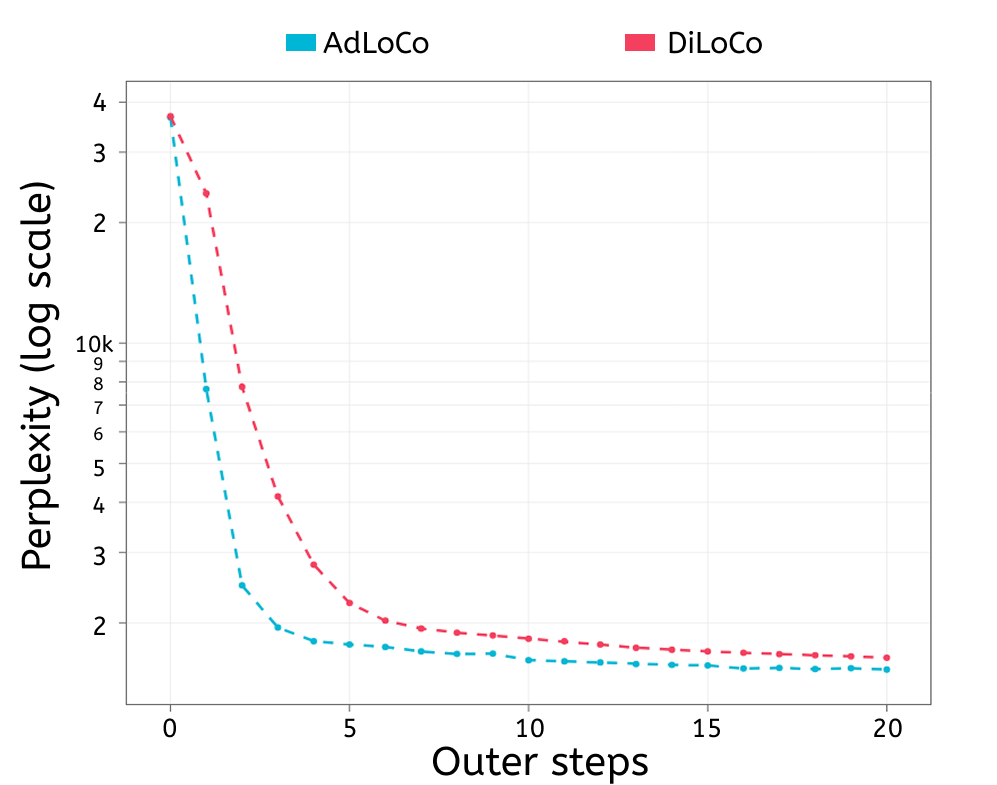}
  \caption{AdLoCo vs DiLoCo comparison}
  \label{fig:1}
\end{minipage}%
\hfill
\begin{minipage}{0.48\textwidth}
  \centering
  \includegraphics[width=\linewidth]{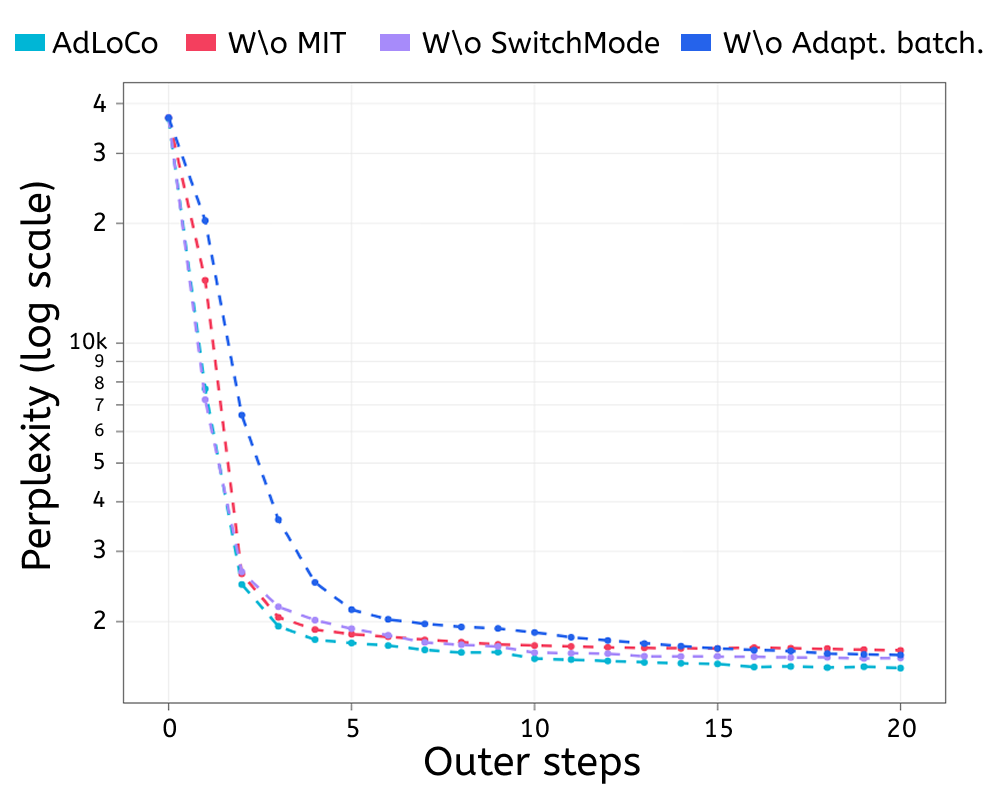}
  \caption{Ablation study}
  \label{fig:2}
\end{minipage}
\end{figure}

Figure \ref{fig:1} highlights the clear performance gap between AdLoCo and DiLoCo in terms of convergence speed and communication efficiency. While DiLoCo already reduces synchronization costs compared to LocalSGD, its fixed batching policy often leads to suboptimal GPU utilization under dynamic workloads. In contrast, AdLoCo leverages adaptive batching to continuously balance computation and communication. As a result, the curves in Figure 1 show that AdLoCo reaches lower perplexity values in fewer training steps, and achieves the target performance substantially faster. This improvement confirms that the integration of adaptive batching, trainer-merger strategies and policy switching allows AdLoCo to better exploit hardware resources than DiLoCo.

\subsection{Ablation study}

To better understand the contribution of each component of \textsc{AdLoCo}, we conducted a series of ablation experiments, the results of which are summarized in Figure~\ref{fig:2}.

\paragraph{Effect of adaptive batching.} 
We first isolate the impact of adaptive batch sizing by comparing \textsc{AdLoCo} with and without the norm-test based adaptive batch scheduler. Without adaptivity, the system struggles with GPU underutilization and suffers from slower convergence. The adaptive variant shows a consistently faster descent in perplexity, highlighting the importance of dynamic batch control for balancing computation and communication.

\paragraph{Effect of trainer merger.} 
Next, we evaluate the role of the trainer merger mechanism. Removing the merger leads to wasted computation from weaker trainers, which slows down convergence. In contrast, the full \textsc{AdLoCo} setup consolidates trainers effectively, improving stability and efficiency.

\paragraph{Effect of gradient accumulation (policy switching).} 
Finally, we assess the impact of enabling gradient accumulation once adaptive batch sizes exceed hardware-friendly limits. Without this switching mechanism, training becomes unstable at large batch regimes. Introducing accumulation restores smooth convergence and prevents memory bottlenecks.

\paragraph{Summary.} 
Across all ablations, each component contributes meaningfully to the overall performance gains. Adaptive batching improves hardware utilization, the trainer merger reduces wasted effort, and gradient accumulation stabilizes training at scale. Together, these design choices explain the superior convergence efficiency of AdLoCo compared to DiLoCo and its ablated variants.

\section{Concluding remarks}
In this work, we introduced AdLoCo, an adaptive extension of DiLoCo, designed to improve the efficiency of distributed training for Large Language Models. By integrating adaptive batching and trainer merging policies, AdLoCo achieves better utilization of computational resources, reduced communication overhead, and more stable convergence. Our theoretical analysis provided guarantees on both expected batch size growth and communication complexity, while empirical results demonstrated tangible improvements over baseline methods in terms of convergence speed and efficiency.

In general, AdLoCo advances the state of distributed optimization by combining the strengths of LocalSGD, adaptive batching, and coordinated trainer merger policies. These contributions make it a promising approach for large-scale model training in resource-constrained environments. 

\section{Future work}
Future work may explore integrating more sophisticated variance-reduction techniques into AdLoCo and applying it at larger scales to further validate its performance. Furthermore, we aim to provide a theoretical justification for phase switching and try to get an accelerated version of the method.

\bibliography{icomp2024_conference}
\bibliographystyle{icomp2024_conference}

\subsection*{Acknowledgements}
The study has been supported by the Ministry of Economic Development of the Russian Federation (agreement No. 139-10-2025-034 dd. 19.06.2025, IGK 000000C313925P4D0002).

\appendix

\section{Appendix}

\subsection{AdLoCo guarantees}
\subsubsection{Lemma 1 (LocalSGD batch growth in non-convex case)}

Following \cite{localsgd}, we make the following assumptions:

The global objective $F$ is $L$-smooth:
\[
\begin{aligned}
F(x') &\le F(x) + \langle \nabla F(x), x' - x\rangle
  + \tfrac{L}{2}\|x' - x\|^2,
\qquad \forall x,x' \in \R^d .
\end{aligned}
\]

There exist constants $\sigma^2, G^2$ such that for all $x$,
\[
  \sigma^2 \le \E_{\xi \sim \mathcal{D}_i}\bigl\| \nabla f(x; \xi) - \nabla F_i(x) \bigr\|^2 \le G^2,
  \qquad
\]

Trainers synchronize (average their models) at iterations $\mathcal I_T \subseteq \{1,\dots,T\}$, with bounded synchronization interval
\[
  \max_{t\in[T]} \bigl( t - \max\{ s\in\mathcal I_T : s \le t\} \bigr) \;\le H.
\]
The stepsize should be bounded as follows:  $\eta_k \le \tfrac{1}{4L}$. \newline

Under these assumptions, the following equation holds for LocalSGD with norm test adaptive batching: 
\begin{equation}
\mathbb{E}[b_k] = \Omega \left(\frac{k\sigma^2}{\eta^2 L(HM+\eta^2)(F(x_0) - F(x^*))}\right)
\end{equation}
\begin{proof}
\begin{equation}
    \mathbb{E}[b_{k+1}]=\mathbb{E}[\mathbb{E}[b_{k+1}|x_k]] \geq \mathbb{E}[\mathbb{E}[\frac{||\nabla \hat{F}(x_k) - \nabla F(x_k)||^2}{\eta^2||\nabla F(x_k)||^2} |x_k]] \ge \mathbb{E}[\frac{\sigma^2}{\eta^2 ||\nabla F(x_k)||^2}] 
\end{equation}
Using Jensen inequality for $\frac{1}{x}$
\begin{equation}
    \mathbb{E}[b_{k+1}] \geq (\frac{\sigma}{\eta})^2 \frac{1}{\mathbb{E}[||\nabla F(x_k)||^2]}
\end{equation}
From \cite{batched_localsgd}, we know that in non-convex case 
\begin{equation}
\mathbb{E}[||\nabla F(x_{out})||^2] = \mathcal{O}(\frac{L(HM+\eta^2)(F(x_0)-F(x^*)}{K})
\end{equation}
Hence $\forall k$ $b_{k+1} \geq b_{k}$, 
\begin{equation}
\mathbb{E}[b_{k+1}]=\max_{t=[0, 1, ..., k]}\mathbb{E}[b_t] \geq \max_{t=[0, 1, ..., k]}(\frac{\sigma}{\eta})^2 \frac{1}{\mathbb{E}[||\nabla F(x_t)||^2]} \geq  (\frac{\sigma}{\eta})^2 \frac{1}{\mathbb{E}[||\nabla F(x_{out,k})||^2]}
\end{equation}
Substitution ends proof. In particular, as will be shown below, when H=1 and M=1, the estimate coincides with classical SGD with adaptive batching.
\end{proof}
\subsubsection{Lemma 2 (SGD convergence with norm test adaptive batching)}
\begin{proof}
Proof is similar to that in \cite{norm_test}, we state it here for convenience and completeness.
Using L-smoothness, we state that 
\begin{equation}
    F(x_{k+1})\leq F(x_{k})-\alpha_k \left<\nabla F(x_k), \nabla F_{B_k}(x_k) \right> + \frac{L\alpha_k^2}{2}||\nabla F_{B_k}(x_k)||^2
\end{equation}
Taking conditional expectation on $x_k$, we have
\begin{equation}
    \mathbb{E}[F(x_{k+1}) | x_k]\leq F(x_{k})-\alpha_k ||\nabla F(x_k)||^2 + \frac{L\alpha_k^2}{2}\mathbb{E}[||\nabla F_{B_k}(x_k)||^2 | x_k]
\end{equation}
Using E-SG, we estimate the second moment:
\begin{equation}
    \mathbb{E}[||\nabla F_{B_k}(x_k)||^2 | x_k] \leq ||\nabla F(x_k)||^2 + E[||\nabla F_{B_k}(x_k) - \nabla F(x_k)||^2] \leq (1+\eta^2)|| \nabla F(x_k)||^2
\end{equation}
Combining previous equations, we get:
\begin{equation}
    \mathbb{E}[F(x_{k+1}) | x_k] \leq F(x_k) + (\frac{(1+\eta^2)L\alpha_k^2}{2} -\alpha_k)||\nabla F(x_k)||^2
\end{equation}
Using full expectation: 
\begin{equation}
\alpha_k(1 - \frac{(1+\eta^2)L\alpha_k}{2}) \mathbb{E}[||\nabla F(x_k)||^2] \leq \mathbb{E}[F(x_k) -F(x_{k+1})]
\end{equation}
\begin{equation}
\sum_{t=0}^{K}\alpha_t(1 - \frac{(1+\eta^2)L\alpha_t}{2}) \mathbb{E}[||\nabla F(x_t)||^2] \leq \mathbb{E}[F(x_0) -F(x_{K+1})] \leq F(x_0)-F(x^*)
\end{equation}
For $\alpha_t = \alpha = \frac{1}{(1+\eta^2)L}$ implies
\begin{equation}
    \frac{1}{K}\sum_{t=0}^{K}\mathbb{E}[||\nabla F(x_{t})||^2] \leq \frac{L(1 + \eta^2)(F(x_0)-F(x^*))}{K}
\end{equation}
\begin{equation}
    \mathbb{E}[||\nabla F(x_{out})||^2] \leq \frac{L(1 + \eta^2)(F(x_0)-F(x^*))}{K}
\end{equation}
where $x_{out}$ is sampled uniformly from $x_t, t\in[0,...,K]$
\end{proof}
\subsubsection{Lemma 3 (Communications amount at AdLoCo)}
\begin{proof}
We can rewrite communication amount of AdLoCo procedure in such manner: $C(N)=\sum_{k=0}^{N} \frac{b_{max}}{b_k}$, so, applying Jensen's inequality for $\frac{1}{x}$ and using Lemma 2:
    \begin{equation}
        \mathbb{E}[C(N)]=b_{max}\sum_{k=0}^{N}\mathbb{E}[\frac{1}{b_k}] \geq b_{max}\sum_{k=0}^{N}\frac{1}{\mathbb{E}[b_k]} 
    \end{equation}
    \begin{equation}
        \mathbb{E}[C(N)] = \mathcal{O}\left(\frac{\eta^2L(1 + \eta^2)(F(x_0)-F(x^*))}{\sigma^2} \sum_{k=0}^N\frac{b_{max}}{k}\right)
    \end{equation}
    \begin{equation}
        \mathbb{E}[C(N)] = \mathcal{O}\left(\frac{b_{max}\eta^2L(1 + \eta^2)(F(x_0)-F(x^*))}{\sigma^2}\ln{N}\right)
    \end{equation}
\end{proof}

\vspace{100mm}
\renewcommand{\arraystretch}{1.1}

\subsection{Hyperparameters}
\begin{table}[h]
\centering
\begin{tabular}{l|l}
\hline
\textbf{Parameter} & \textbf{Value} \\ \hline
num\_outer\_steps & 20 \\ 
num\_inner\_steps & 200 \\ 
lr\_inner & 2e-5 \\ 
lr\_outer & 0.5 \\ 
nodes\_per\_gpu & 4 \\ 
num\_init\_trainers & 4 \\ 
initial\_batch\_size & 1 \\ 
merge\_frequency & 3 \\ 
$\eta$ & 0.8 \\ 
$\vartheta$ & 0.01 \\ 
$\nu$ & 0.3 \\ \hline
\end{tabular}
\caption{Key hyperparameters used for training}
\end{table}

\label{algo}
\subsection{AdLoCo algorithm}

\begin{algorithm}[H]
\caption{\textsc{CheckMerge}: select $w$ worst trainers by requested batch}
\label{alg:checkmerge}
\begin{algorithmic}[1]
\STATE \textbf{Input:} requested batches $\{b^{\mathrm{req}}_i\}_{i=1}^{k}$
\STATE \textbf{Input:} merge hyperparameter $w$
\IF{$w=0$ \OR $k \le 1$}
  \STATE \textbf{return} $\varnothing$
\ENDIF
\STATE sort trainers increasingly by $b^{\mathrm{req}}_i$ to get order $(i_{(1)},\ldots,i_{(k)})$
\IF{$w \le k$}
  \STATE \textbf{return} $S \gets \{\,i_{(1)},\ldots,i_{(w)}\,\}$ \COMMENT{``$w$ worst'' policy}
\ELSE
  \STATE \textbf{return} $\varnothing$
\ENDIF
\end{algorithmic}
\end{algorithm}

\begin{algorithm}[H]
\caption{\textsc{DoMerge}: weighted parameter merge of trainers in $S$}
\label{alg:domerge}
\begin{algorithmic}[1]
\STATE \textbf{Input:} set $S$ of trainers to merge
\STATE \textbf{Input:} models $\{x_j^{(t,H)}\}_{j\in S}$
\STATE \textbf{Input:} weights $\{b^{\mathrm{req}}_j\}_{j\in S}$
\STATE $w_{\text{sum}} \gets \sum_{j\in S} b^{\mathrm{req}}_j$
\STATE $x_{\text{merge}} \gets w_{\text{sum}}^{-1}\!\sum_{j\in S} b^{\mathrm{req}}_j \, x_j^{(t,H)}$
\STATE pick representative $r \in S$ with $\max b^{\mathrm{req}}_j$
\STATE $x_r^{(t,H)} \gets x_{\text{merge}}$
\STATE remove $S\setminus\{r\}$ from the trainer set; \; update $k \gets k - (|S|-1)$
\STATE carry optimizer state of $r$ forward
\end{algorithmic}
\end{algorithm}

\begin{algorithm}
\caption{AdLoCo: Adaptive Batching + Merging + Switch mode (DiLoCo core)}
\label{alg:adloco}
\begin{algorithmic}[1]
\STATE \textbf{Input:} number of trainers $k$
\STATE \textbf{Input:} number of workers $M$
\STATE \textbf{Input:} initial model $x^{(0)}$
\STATE \textbf{Input:} data shards $\{\mathcal{D}_i\}_{i=1}^k$
\STATE \textbf{Input:} inner steps $H$
\STATE \textbf{Input:} optimizers \textsc{InnerOpt}, \textsc{OuterOpt}
\STATE \textbf{Input:} adaptive batching params $(\eta,\vartheta)$
\STATE \textbf{Input:} max batch $b_{\max}$, switch multiplier $n{=}2$
\STATE \textbf{Input:} merge hyperparameter $w$ \; (merge $w$ worst trainers by requested batch)
\FOR{outer step $t=1\ldots T$}
  \IF{$k>1$}
    \STATE $\mathbb{M}_t \gets \textsc{CheckMerge}\big(\{b^{\mathrm{req}}_i\}_{i=1}^{k},\, w\big)$
    \IF{$\mathbb{M}_t \neq \varnothing$}
      \STATE \textsc{DoMerge}$(\mathbb{M}_t)$ 
    \ENDIF
  \ENDIF

  \IF{$b^{\mathrm{req}}_{i} > n \cdot b_{\max}$}
  
    \STATE $\text{accum} \gets \left\lceil b_i^{\mathrm{req}}/b_{\max}\right\rceil$, \quad $b^{\mathrm{micro}} \gets b_{\max}$
    \FOR{$m=1\ldots \text{M}$}
    \FOR{$s=1\ldots \text{accum}$}
      \STATE sample $B \subset \mathcal{D}_m$ with $|B|=b^{\mathrm{micro}}$
      \STATE $g \gets g + \nabla \mathcal{L}(x_m^{t-1};B)$
    \ENDFOR
    \STATE $x_m^{t} \gets \textsc{InnerOpt}(x_m^{(t-1)}, g)$
    \ENDFOR

\ELSE
  \STATE $b^{\mathrm{micro}} \gets \min\!\big(b^{\mathrm{req}}_{i},\, b_{\max}\big)$
  \quad
  \FOR{$T_i \in \mathbb{T}_t$}
  \FOR{$m \in T_i$}
        \STATE $x_{m}^{(t,0)} \gets x_{T_{i}}^{(t-1)}$
        \STATE compute requested batch $b^{\mathrm{req}}_{T_i}$ via \emph{norm test} (see Related work / Norm test)
        
        \STATE store $b^{\mathrm{req}}_{T_i}$ for the next outer step
      
  \FOR{inner step $h=1\ldots H$}
    \STATE $g \gets \nabla \mathcal{L}(x_m^{t-1};B)$
    \STATE $x_m^{(t,h)} \gets \textsc{InnerOpt}(x_m^{(t,h-1)}, g)$
  \ENDFOR
  \ENDFOR
  \ENDFOR
\ENDIF
\FOR{$T_i\in \mathbb{T}_t$}
\STATE $M_i \gets |T_i|$
\STATE $\Delta_{T_i}^{(t)} \gets \frac{1}{M}\sum_{m\in T_i}\big(x^{(t-1)} - x_m^{(t,H)}\big)$
\STATE $x_{T_i}^{(t)} \gets \textsc{OuterOpt}(x_{T_i}^{(t-1)}, \Delta_i^{(t)})$
\ENDFOR
\ENDFOR
\end{algorithmic}
\end{algorithm}

\end{document}